\documentclass[letterpaper]{article}
\usepackage[letterpaper]{geometry}

\usepackage[utf8]{inputenc}
\usepackage[T1]{fontenc} 
\usepackage[cjk]{kotex}

\usepackage{amsmath}
\usepackage{graphicx}
\usepackage{url}
\usepackage{xcolor}
\usepackage{latexsym}

\usepackage{hyperref}
\definecolor{darkblue}{rgb}{0, 0, 0.5}
\hypersetup{colorlinks=true,citecolor=darkblue, linkcolor=darkblue, urlcolor=darkblue}
\usepackage{tabularx}
\usepackage{datetime}
\usepackage{arydshln}

\usepackage{setspace}
\usepackage{lineno}

\usepackage{booktabs}

\usepackage[authoryear,round]{natbib}
\usepackage{algorithmic,algorithm}
\renewcommand{\algorithmiccomment}[1]{\bgroup\hfill$\triangleright$~#1\egroup}

\usepackage[linguistics]{forest} 
\usepackage{synttree}
\usepackage{subcaption}
\usepackage{tikz-dependency}

\usepackage{gb4e}

\title{\textbf{{
Annotating Korean adnominal ending constructions in corpus data: Beyond relative-clause identification
}}}

\author{
Jungyeul Park$^{1*}$ and Chulwoo Park$^{2}$\thanks{Jungyeul Park and Chulwoo Park are corresponding authors.}\\
$^{1}$KAIST, South Korea. {\tt jungyeul@kaist.ac.kr} \\
$^{2}$Anyang University, South Korea. {\tt cwpa@anyang.ac.kr}
}
\date{ 
}

\begin{document}

\maketitle

\begin{abstract}
The Korean adnominal ending \texttt{ETM} occurs in diverse noun-modifying constructions, including relative-clause-like modifiers, adjectival and copular forms, bound-noun constructions, and lexicalized expressions. This paper argues that \texttt{ETM} is not a direct marker of relative-clause structure, but a morphological exponent shared by several adnominal constructions. We propose a corpus-based typology that distinguishes these constructions using predicate type, auxiliary structure, argument-structural compatibility, head-noun restriction, and lexicalized patterns. We operationalize the typology as a construction-sensitive annotation layer for the KLUE dependency treebank, implemented through an ordered rule-based procedure and evaluated by manual validation. Productive relative-clause-like uses account for 39.4\% of the analyzed instances; the remainder consists mainly of adjectival, copular, bound-nominal, modal, temporal, and collocational constructions. The findings show that Korean relative-clause-like modification cannot be identified from adnominal morphology alone.
\end{abstract}


\doublespacing

\section{Introduction}

Korean adnominal modification cannot be reduced to relative clause formation. The Korean adnominal ending \texttt{ETM} appears in relative-clause-like modifiers, but it also occurs in adjectival modification, copular adnominal forms, bound-noun constructions, and lexicalized or constructional collocations.\footnote{\texttt{ETM} is the Sejong corpus tag for \textit{\textbf{e}omi} (`verbal ending') functioning as a \textbf{t}erminal \textbf{m}odifier. The Sejong scheme takes the \textit{eojeol}, the Korean spacing unit, as the basic orthographic unit of analysis; an \textit{eojeol} may include lexical morphemes, grammatical endings, particles, and punctuation, while tags such as \texttt{ETM} mark the morphemic elements inside that unit.} The central issue is therefore not simply that \texttt{ETM} is morphologically ambiguous, but that the same adnominal ending serves as the overt exponent of several distinct noun-modifying constructions. Treating \texttt{ETM} itself as a marker of relative clause structure obscures this constructional heterogeneity and makes it difficult to compare Korean adnominal modification across corpora and annotation schemes. For corpus-based research, this creates an annotation problem as well as a descriptive one: formally identical adnominal morphology must be separated into constructionally distinct uses before its distribution can be interpreted.

This problem is especially visible in treebank annotation, where constructional distinctions are often flattened into coarse syntactic labels. In Sejong-style treebanks, including the KLUE dependency treebank \citep{park-etal-2021-klue}, adnominally marked predicates are commonly grouped under broad modifier labels such as \texttt{vp\_mod}. Such labels identify where modification occurs, but they do not distinguish whether a given \texttt{ETM} form realizes a relative-clause-like modifier, an adjectival modifier, a copular adnominal form, a bound-noun construction, or a collocational expression. Korean Universal Dependencies adopts a more differentiated strategy, distributing \texttt{ETM}-related constructions across relations such as \texttt{acl:relcl}, \texttt{amod}, \texttt{advcl}, \texttt{flat}, \texttt{conj}, and \texttt{dep} \citep{de-marneffe-etal-2014-universal,mcdonald-etal-2013-universal,chun-EtAl:2018:LREC}. However, previous work has raised concerns about the linguistic reliability of Korean UD annotation and its mapping of Korean-specific constructions onto the crosslinguistic UD relation inventory \citep{noh-etal-2018-enhancing,seo-etal-2019-ud,lee-oh-kim-2019-ud,han-etal-2020-annotation,jo-etal-2023-k,kim-etal-2024-kud}. Thus, neither the coarse Sejong-style treatment nor the existing distribution of dependency relations in Korean UD provides a transparent constructional classification of \texttt{ETM}.

This paper develops a constructional typology of Korean \texttt{ETM} uses and applies it to the KLUE dependency treebank. The goal is not to identify relative clauses by searching for \texttt{ETM}, but to determine which adnominal construction a given \texttt{ETM} instance realizes. We therefore distinguish productive relative-clause-like modifiers from adjectival, copular, bound-nominal, modal, temporal, and collocational constructions that share the same adnominal morphology. In this respect, the present study follows a broader corpus-linguistic tradition in which recurrent form--meaning pairings are treated as constructional patterns rather than as simple surface strings \citep{gries-stefanowitsch-2004-extending}. We operationalize the typology as a construction-sensitive annotation layer, using an ordered rule-based procedure and evaluating the resulting labels through manual validation. Although our task differs from general syntactic-complexity measurement, it shares with previous work on automatic syntactic analysis the need to make computationally derived categories explicit and empirically validated \citep{lu-2010-automatic}. The resulting analysis treats \texttt{ETM} not as a construction type in itself, but as a morphological form whose interpretation depends on predicate type, auxiliary structure, head-noun restrictions, and lexicalized constructional patterns.

The contribution of the paper is threefold. First, it proposes a linguistically motivated typology of Korean \texttt{ETM} constructions that separates relative-clause-like uses from other adnominal constructions. Second, it operationalizes this typology over the KLUE dependency treebank through an ordered rule-based annotation procedure and validates the resulting labels on a manually checked sample. Third, it presents a corpus analysis showing that relative-clause-like uses constitute only one part of the empirical distribution of \texttt{ETM}. The broader conclusion is that Korean relative-clause analysis must be construction-sensitive: adnominal morphology alone is not enough to identify relative clause structure. We release the label schema,  documentation, and the entire \texttt{ETM}-annotated training dataset derived from KLUE.\footnote{For anonymous review, these materials are distributed through an anonymized repository: \url{https://anonymous.4open.science/r/annotating-korean-etm-D844/}. The repository includes the annotation code, label schema, documentation, the entire \texttt{ETM}-annotated training dataset, and supplementary materials with category definitions and representative corpus examples. Since KLUE is distributed under the Creative Commons Attribution--ShareAlike 4.0 International License, our release will follow the same licensing conditions.}

\section{Background}

\subsection{Korean adnominal modification and linguistic diagnostics}

Korean noun modification includes a range of adnominal constructions that share surface morphology but differ in syntactic and interpretive organization. Some \texttt{ETM} forms introduce predicate-based nominal modification, as in canonical relative-clause-like expressions, while others function as adjectival modifiers, copular modifiers, bound-noun constructions, or constructionally restricted nominal modifiers. Korean also includes recurrent patterns such as \texttt{Verb+ETM} 때 \textit{ttae} (`time'), modal expressions involving 수 \textit{su} (`possibility/way') and 있다/없다 \textit{itda/eopda} (`exist/not exist'), and collocationalized sequences in which the adnominal form is part of a semi-fixed expression rather than an ordinary productive relative clause. A construction-sensitive account must therefore identify the syntactic and lexical conditions under which an \texttt{ETM} form participates in each construction type.

The linguistic literature has long noted that Korean adnominal clauses differ in important ways from relative clause systems familiar from languages such as English. \citet{jaeil-2015-gapless} observe that Korean relative clauses do not permit mood markers, whereas clauses denoting propositions, facts, or other content-like meanings require them. \citet{taein-2022-adnominal} relate this contrast to the distinction between finite and non-finite clause types, noting that noun-modifying clauses in Korean are characteristically non-finite. In a related typological perspective, \citet{jaehoon-2012-functional-typological} argue that Korean is better analyzed in terms of attributive clause constructions than in terms of relative clauses defined by overt gaps or relative pronouns. These studies motivate the view that Korean adnominal modification should not be classified by direct analogy with relative-clause systems in languages with different morphosyntactic resources.

Previous studies of Korean adnominal endings have also identified semantic and morphosyntactic factors relevant to the classification pursued here. \citet{park-2009-meanings} analyzes -은/ㄴ \textit{-(u)n} and -을/ㄹ \textit{-(u)l} in terms of realis and irrealis modality, while \citet{takachi-2008-essay} distinguishes established, certain, and undetermined domains associated with -은/ㄴ \textit{-(u)n}, -는 \textit{-nun}, and -을/ㄹ \textit{-(u)l}, respectively. \citet{kim-2016-functional} relates the interpretation of adnominal endings to the ordering of the adnominal-clause event, the matrix event, and the utterance time. These analyses show that Korean adnominal endings contribute temporal, aspectual, and modal information and therefore cannot be treated as semantically empty markers of relative-clause formation.

The same line of work points to diagnostics that are directly useful for corpus annotation. \citet{kim-2016-functional} shows that the distribution and interpretation of adnominal endings depend on predicate type: verbal predicates contrast with adjectival and copular predicates in their compatibility with particular endings. This predicate-type sensitivity provides a basis for separating verbal relative-clause-like modifiers from adjectival and copular adnominal constructions. \citet{takachi-2008-essay} further shows that interpretations such as possibility, obligation, ability, and volition arise in combination with particular head nouns, including nouns meaning `opportunity', `necessity', `intention', and `ability'. Such cases demonstrate that the interpretation of an adnominal expression may be determined jointly by the ending and a lexically restricted nominal head, as in the bound-noun and construction-specific patterns examined in this study.

Argument-structural diagnostics provide another relevant dimension. \citet{park-2002-criteria} uses relativization behavior to distinguish selected complements from adjuncts, observing that a complement may correspond to the head of a relative clause whereas an adjunct generally may not. \citet{younghee-2004-criterion} proposes particle omission as a related diagnostic: complements may occur without an overt case particle under conditions in which adjuncts resist such omission. These diagnostics help determine whether the modified nominal can be interpreted as an argument associated with the adnominal predicate, supporting a relative-clause-like analysis, or whether the modifier--head relation is better attributed to another construction.

These studies provide several diagnostics for classifying Korean adnominal constructions: predicate type, finiteness and mood compatibility, semantic relation between the predicate and the nominal head, head-noun restriction, lexicalization, and argument-structural compatibility. These dimensions have generally been examined separately in studies of adnominal-ending semantics, modality, finiteness, or argumenthood. The present typology integrates them into a corpus-oriented classification that systematically separates productive relative-clause-like modification from adjectival, copular, bound-noun, and lexically restricted adnominal constructions.

\subsection{\texttt{ETM} in Korean treebanks}

Existing Korean treebanks represent adnominal modification at different levels of granularity. In Sejong-style resources, including the KLUE dependency treebank \citep{park-etal-2021-klue}, adnominally marked predicates are typically represented with broad modifier labels such as \texttt{vp\_mod}. This representation is useful for identifying modifier dependencies, but it does not encode the constructional subtype of the adnominal expression. It therefore records the dependency configuration while leaving the relevant constructional distinction implicit.

Korean treebanks in Universal Dependencies adopt a different strategy. \texttt{ETM}-related constructions may be distributed across dependency relations such as \texttt{acl:relcl}, \texttt{amod}, \texttt{advcl}, \texttt{flat}, \texttt{conj}, and \texttt{dep} \citep{de-marneffe-etal-2014-universal,mcdonald-etal-2013-universal,chun-EtAl:2018:LREC}. This creates a more differentiated annotation space, but the analysis depends on how consistently Korean morphosyntactic structures are mapped onto the crosslinguistic UD relation inventory. Previous work has raised concerns about Korean UD annotation and its treatment of Korean-specific constructions \citep{noh-etal-2018-enhancing,seo-etal-2019-ud,lee-oh-kim-2019-ud,han-etal-2020-annotation,jo-etal-2023-k,kim-etal-2024-kud}.

The issue for the present paper is not whether one treebank scheme is preferable to another, but that neither scheme directly provides the constructional classification needed for the analysis of \texttt{ETM}. Sejong-style annotation underdistinguishes adnominal construction types, while UD-style annotation distributes them across dependency relations whose correspondence to Korean constructional categories must be interpreted with care. We therefore treat existing treebank annotation as the structural input to a separate construction-sensitive layer, rather than as the final classification itself.

\section{Data} \label{sec:data}

\subsection{KLUE dependency treebank}

The main corpus used in this study is the KLUE dependency treebank \citep{park-etal-2021-klue}, a morphologically and syntactically annotated Korean dependency treebank following the Sejong style of annotation. {We use the training split of KLUE as the empirical basis for examining the constructional distribution of \texttt{ETM}.} Our analysis targets all instances in this split in which the morphological analysis contains the adnominal ending \texttt{ETM}. {The treebank is therefore not treated merely as a source of dependency labels, but as a corpus in which formally identical adnominal morphology can be examined across different syntactic and lexical environments.}

\begin{figure}[!ht]
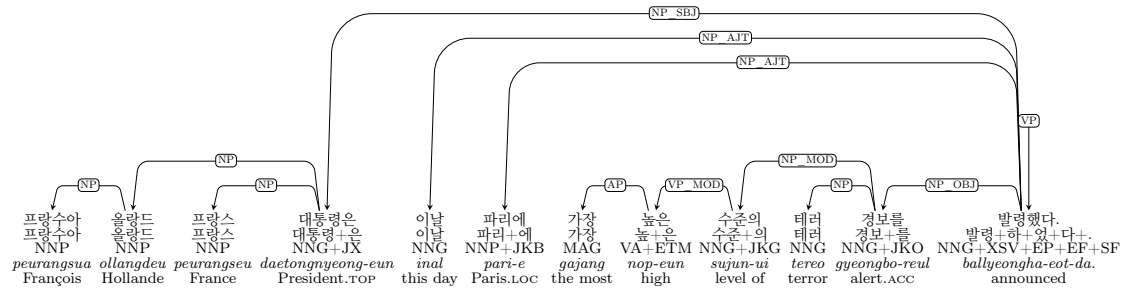

\begin{center}
{\footnotesize
\resizebox{.99\textwidth}{!}
{\footnotesize
\begin{dependency}
\begin{deptext}
프랑수아 \& 올랑드 \& 프랑스 \& 대통령은 \& 이날 \& 파리에 \& 가장 \& 높은 \& 수준의 \& 테러 \& 경보를 \& 발령했다. \\
프랑수아 \& 올랑드 \& 프랑스 \& 대통령+은 \& 이날 \& 파리+에 \& 가장 \& 높+은 \& 수준+의 \& 테러 \& 경보+를 \& 발령+하+었+다+. \\
NNP \& NNP \& NNP \& NNG+JX \& NNG \& NNP+JKB \& MAG \& VA+ETM \& NNG+JKG \& NNG \& NNG+JKO \& NNG+XSV+EP+EF+SF \\
\textit{peurangsua} \& \textit{ollangdeu} \& \textit{peurangseu} \& 
\textit{daetongnyeong-eun} \& \textit{inal} \& \textit{pari-e} \& 
\textit{gajang} \& \textit{nop-eun} \& \textit{sujun-ui} \& 
\textit{tereo} \& \textit{gyeongbo-reul} \& 
\textit{ballyeongha-eot-da.} \\
{François} \& {Hollande} \& {France} \& {President.\textsc{top}} \& {this day} \& {Paris.\textsc{loc}} \& {the most} \& {high} \& {level of} \& {terror} \& {alert.\textsc{acc}} \& {announced} \\
\end{deptext}
\depedge{2}{1}{NP}
\depedge{4}{2}{NP}
\depedge{4}{3}{NP}
\depedge{12}{4}{NP\_SBJ}
\depedge{12}{5}{NP\_AJT}
\depedge{12}{6}{NP\_AJT}
\depedge{8}{7}{AP}
\depedge{9}{8}{VP\_MOD}
\depedge{11}{9}{NP\_MOD}
\depedge{11}{10}{NP}
\depedge{12}{11}{NP\_OBJ}
\deproot{12}{VP}
\end{dependency}
}}
\end{center}
    \caption{An example from the KLUE dependency treebank (sentence meaning: `François Hollande, the President of France, announced the highest level of a terror alert in Paris on this day.').}
    \label{klue-ex-dep}
\end{figure}

The basic tokenization unit in KLUE is the \textit{eojeol}. As a result, punctuation is not always separated from the preceding word, and a single token may contain multiple morphemes. This is visible in Figure~\ref{klue-ex-dep}, where the final \textit{eojeol} 발령했다. \textit{ballyeonghaetda.} (`announced') consists of the morpheme sequence 발령+하+었+다+. \textit{ballyeong+ha+eoss+da+.} and includes sentence-final punctuation within the same \textit{eojeol}. Dependency relations are annotated over these \textit{eojeol}-level units, and the treebank follows the head-final orientation characteristic of Korean. In addition, root labels are not represented by a single uniform \texttt{root} category, but by root categories that retain information about the syntactic type of the predicate, such as \texttt{VP} or \texttt{VNP}.

For the purposes of the present typology, we make direct use of the morphological POS information provided in KLUE. The categories most relevant to classification include \texttt{ETM} for adnominal endings, \texttt{EC} for connective endings, \texttt{NNB} for bound nouns, \texttt{VX} for auxiliary verbs, and \texttt{VCP} for the copula. Predicate-related evidence are identified mainly through \texttt{VV}, \texttt{VA}, and verbal and adjectival derivational suffixes (\texttt{XSV} and \texttt{XSA}), while nominal elements are identified through tags such as \texttt{NNG}, \texttt{NNP}, and the nominalizing ending \texttt{ETN}. These tags provide the local evidence needed to distinguish, for example, verbal relative-clause-like uses from adjectival, copular, bound-nominal, and auxiliary-based adnominal constructions.

\texttt{ETM} instances were extracted automatically from the morphological layer of the treebank by selecting all \textit{eojeol}s whose morpheme sequence contains \texttt{ETM}. The resulting set was then used as the annotation target for the typology proposed in this paper. {Because the relevant distinction is constructional rather than purely morphological, each extracted instance was analyzed together with its morpheme sequence, neighboring lexical material, head noun, and local dependency configuration.} {This context is necessary because the same \texttt{ETM} ending may participate in different adnominal constructions depending on predicate type, auxiliary structure, and the lexical or constructional status of the modified noun.}

\subsection{Auxiliary lexical resource}

As an auxiliary resource for annotation support, we consulted the Sejong lexical resource distributed by the National Institute of Korean Language. The resource was not used to determine the basic inventory of \texttt{ETM} instances, which comes entirely from KLUE, but to support decisions in cases where the interpretation of an \texttt{ETM} construction depends on lexical properties of the predicate. In particular, we consulted lexical entries when annotation required information about predicate class, subcategorization, or argument structure that was not fully recoverable from the local treebank context alone.

For this purpose, the most relevant fields in the lexical entries are the predicate lemma, the subcategorization frame, and example based information about argument realization. These fields help disambiguate difficult cases in which a surface \texttt{ETM} sequence may support more than one structural interpretation. {The lexical resource is therefore used only as supporting evidence for constructional classification, not as an independent source of labels.}

Figure~\ref{sjdict-ballyeonghada} shows a compact example from the lexical entry for 발령하다 \textit{ballyeonghada} (`to announce'). The entry specifies the verbal lemma together with a transitive frame and associated argument information. In the present study, such information is used only insofar as it helps determine whether a given \texttt{ETM} construction is compatible with a relative-clause-like interpretation or better analyzed as another construction type.

\begin{figure}[!ht]
    \centering
\footnotesize
\begin{verbatim}
<orth>발령하다</orth>  <!-- issue; announce -->
<entry n="1" pos="vv">
  <sense>
    <frame_grp type="FTR">
      <frame>X=N0-이 Y=N1-을 V</frame>
      <subsense>
        <sel_rst arg="X" tht="AGT">인간|인간집단</sel_rst>  <!-- human | human group -->
        <sel_rst arg="Y" tht="THM">(경보|호우주의보)</sel_rst>  <!-- alarm | heavy-rain advisory -->
        <eg>화재 경보를 발령하다.</eg>  <!-- issue a fire alarm -->
      </subsense>
    </frame_grp>
  </sense>
</entry>
\end{verbatim}
    \caption{Compact example from the lexical entry for 발령하다 \textit{ballyeonghada} (`to announce') in the Sejong lexical resource.}
    \label{sjdict-ballyeonghada}
\end{figure}

A fuller description of the XML structure of this lexical resource is not necessary for the main argument of the paper and is therefore omitted here. What matters for the present annotation task is only that the resource provides lemma based lexical information that can be consulted in difficult cases involving predicate interpretation and argument structure.

\section{Typology of \texttt{ETM} constructions} \label{sec:typology}

{This section develops the constructional typology used to analyze \texttt{ETM} instances in the KLUE dependency treebank.} The typology is designed to distinguish formally similar adnominal forms that differ in syntactic function, lexical restriction, and constructional status. {The basic assumption is that \texttt{ETM} is not itself a construction type. Rather, it is an adnominal morphological form that appears across several noun-modifying constructions.} We therefore distinguish productive relative-clause-like uses from adjectival and copular modifiers, bound-noun and modal-temporal constructions, lexicalized or constructionally restricted patterns, and cases that remain unresolved because of annotation error or insufficient structural evidence.

\subsection{Design principles}

The typology is based on four criteria. First, we distinguish constructions according to their \emph{syntactic function}, in particular whether the \texttt{ETM} form participates in a productive relative-clause-like modifier or in another type of adnominal modification. Second, we distinguish \emph{productive} predicate--noun combinations from \emph{lexicalized or constructionally fixed} expressions, since some \texttt{ETM} sequences behave as semi-fixed patterns rather than freely generated relative clauses. Third, we consider whether the modified noun is interpreted as a semantically open head nominal or as a \emph{constructionally restricted head}, such as {때} \textit{ttae} (`time'), {수} \textit{su} (`possibility/way'), or nouns occurring in recurrent collocational expressions. Fourth, we isolate cases that cannot be classified reliably because of annotation inconsistency or insufficient structural evidence.

{On this basis, the typology is organized around the constructional relation between the adnominal form and the following noun.} The first group includes lexicalized or constructionally restricted modifiers, where the \texttt{ETM} form is part of a semi-fixed expression or occurs with a restricted head noun. The second group includes non-relative descriptive modifiers, such as adjectival and copular adnominal forms. The third group includes productive relative-clause-like modifiers, where a predicate or predicate complex modifies an open nominal head. The fourth is a residual \texttt{undefined} class for cases that cannot be analyzed reliably.

\subsection{Category inventory}

Table~\ref{tab:etm-typology} summarizes the inventory used in the annotation.

\begin{table*}[!ht]
\centering
\footnotesize
\begin{tabular}{p{2.8cm} p{4.6cm} p{6.1cm}}
\toprule
\textbf{Label} & \textbf{Structural cue} & \textbf{Interpretation} \\
\midrule
\texttt{colloc+etm}
& lexicalized \texttt{ETM} + noun sequence
& collocational or semi-fixed modifier \\

\texttt{modif+adj+etm}
& \texttt{VA+ETM} + noun
& adjectival modifier \\

\texttt{modif+vcp+etm}
& copular or negative copular adnominal form + noun
& copular descriptive modifier \\

\texttt{modif+ndaneun+etm}
& \texttt{VV} + 는다는 \textit{neundaneun} / ㄴ다는 \textit{ndaneun} + noun
& content-like or quotative nominal modifier \\

\texttt{modif+su-iss+etm}
&  수 \textit{su} (`possibility/way') + 있/없 \textit{iss/eops}+\texttt{ETM} + noun
& modal construction with restricted head noun \\

\texttt{modif+etm+ttae}
& \texttt{Verb+ETM} + 때 \textit{ttae} (`time')
& temporal construction with fixed head noun \\

\texttt{modif+etm+nnb}
& \texttt{Verb+ETM} + bound noun
& bound-noun or semantically light nominal construction \\

\texttt{modif+vx+vx+etm}
& auxiliary complex + \texttt{ETM} + noun
& constructional verbal modifier \\

\texttt{rc+vv+etm}
& \texttt{VV/XSV+ETM} + noun
& canonical relative-clause-like modifier \\

\texttt{rc+vv+vx+etm}
& \texttt{VV+EC+VX+ETM} + noun
& relative-clause-like modifier with auxiliary complex \\

\texttt{rc+va+vx+etm}
& \texttt{VA+EC+VX+ETM} + noun
& adjective-based relative-clause-like modifier \\

\texttt{undefined}
& structurally inconsistent or annotation error
& unresolved case \\
\bottomrule
\end{tabular}

\caption{Inventory of \texttt{ETM} categories used in the annotation.}
\label{tab:etm-typology}
\end{table*}

{The categories in Table~\ref{tab:etm-typology} should be read as constructional types, not simply as surface POS patterns.} The first group, consisting of \texttt{colloc+etm}, \texttt{modif+ndaneun+etm}, \texttt{modif+su-iss+etm}, \texttt{modif+etm+ttae}, and \texttt{modif+vx+vx+etm}, includes cases in which the \texttt{ETM} form appears in a lexicalized or constructionally restricted pattern. These expressions may resemble relative clauses on the surface, but their interpretation is constrained by a fixed expression, a restricted head noun, or a conventionalized construction.

The second group, \texttt{modif+adj+etm} and \texttt{modif+vcp+etm}, includes non-relative descriptive modifiers. In these cases, the \texttt{ETM} form functions primarily as an adjectival or copular nominal modifier rather than as a productive predicate-based relative clause.

The third group, \texttt{rc+vv+etm}, \texttt{rc+vv+vx+etm}, and \texttt{rc+va+vx+etm}, captures productive relative-clause-like modifiers. These are the cases most relevant to the identification of clause-based nominal modification in the corpus. The distinction among the three classes reflects whether the modifier is headed by a simple verbal predicate, a verbal predicate with an auxiliary complex, or an adjective-based complex.

Finally, \texttt{undefined} is used for cases that cannot be assigned a reliable category. This includes annotation errors in the source corpus, incomplete or structurally anomalous forms, and tokens for which the local evidence is insufficient for classification.

\subsection{Decision procedure}

The typology is operationalized by an ordered symbolic procedure, but the ordering reflects linguistic precedence among construction types rather than a purely technical classification strategy. Before the procedure is applied, the local constructional context of each \texttt{ETM} instance is recovered from the dependency structure. The dependency head of the \texttt{ETM}-bearing predicate is treated as its candidate modified nominal, while its immediate dependents provide the local argument or auxiliary context required by particular constructional conditions. Dependency arcs are therefore used to identify the elements participating in the construction; the dependency-relation labels themselves are not used as category-defining criteria.

Some expressions that are formally compatible with a relative-clause-like analysis are classified instead as lexicalized or constructionally restricted modifiers when they match a more specific pattern. In particular, collocational and fixed constructional cases are resolved before productive relative-clause-like categories are considered. This prevents frequent expressions such as -에 대한 \textit{-e daehan} $N$ (`$N$ concerning \ldots'), Verb+ETM 때 \textit{ttae} (`when Verb'), and 수 있다/없다 \textit{su itda/eopda} (`can/cannot') constructions from being overanalyzed as ordinary relative clauses.

Given the dependency-linked predicate, nominal head, and local dependents, the decision procedure first checks whether the \texttt{ETM} form belongs to a listed collocational expression. It then tests for non-relative descriptive classes, including adjectival and copular modifiers, followed by constructionally restricted patterns such as \texttt{modif+ndaneun+etm}, \texttt{modif+su-iss+etm}, \texttt{modif+etm+ttae}, and \texttt{modif+vx+vx+etm}. Only after these classes are excluded are the productive relative-clause-like categories considered. Instances that do not satisfy any category, lack the required dependency-linked context, or contain clear annotation inconsistencies are assigned to \texttt{undefined}.

\begin{algorithm}[!ht]
\caption{Ordered classification of \texttt{ETM} constructions}
\label{algo}
\begin{algorithmic}[1]
\STATE \textbf{Input:} an \texttt{ETM} instance $x$ with its local morphological and syntactic context
\STATE \textbf{Output:} an \texttt{ETM} construction label

\STATE \textbf{function} \textsc{ClassifyETM}$(x)$
\begin{ALC@g}
  \IF{the \textbf{Collocation} condition is met}
    \STATE \textbf{return} \texttt{colloc+etm}

  \ELSIF{the \textbf{Adjective+ETM Noun} condition is met}
    \STATE \textbf{return} \texttt{modif+adj+etm}

  \ELSIF{the \textbf{Copula+ETM Noun} condition is met}
    \STATE \textbf{return} \texttt{modif+vcp+etm}

  \ELSIF{the \textbf{Verb+ETM Noun} condition is met with 는다는/ㄴ다는}
    \STATE \textbf{return} \texttt{modif+ndaneun+etm} \COMMENT{\textit{-neundaneun}/\textit{-ndaneun}: content-like or quotative form}

  \ELSIF{the \textbf{수+있/없+ETM Noun} condition is met}
    \STATE \textbf{return} \texttt{modif+su-iss+etm} \COMMENT{\textit{su} `possibility/way' + \textit{iss/eops} `exist/not exist'}

  \ELSIF{the \textbf{Verb+ETM+때} condition is met}
    \STATE \textbf{return} \texttt{modif+etm+ttae} \COMMENT{\textit{ttae} `time'}

  \ELSIF{the \textbf{Verb+ETM Bound Noun} condition is met}
    \STATE \textbf{return} \texttt{modif+etm+nnb} \COMMENT{e.g., 것 \textit{geos} `thing'}

  \ELSIF{the \textbf{VX+EC+VX+ETM Noun} condition is met}
    \STATE \textbf{return} \texttt{modif+vx+vx+etm}

  \ELSIF{the \textbf{Verb+EC+VX+ETM Noun} condition is met}
    \STATE \textbf{return} \texttt{rc+vv+vx+etm}

  \ELSIF{the \textbf{Adjective+EC+VX+ETM Noun} condition is met}
    \STATE \textbf{return} \texttt{rc+va+vx+etm}

  \ELSIF{the \textbf{Verb+ETM Noun} condition is met}
    \STATE \textbf{return} \texttt{rc+vv+etm}

  \ELSE
    \STATE \textbf{return} \texttt{undefined}
  \ENDIF
\end{ALC@g}
\STATE \textbf{end function}

\end{algorithmic}
\end{algorithm}

Algorithm~\ref{algo} formalizes this ordered procedure. Its ordering is linguistically significant: more specific and constructionally restricted patterns are tested before productive relative-clause-like categories, thereby preventing fixed expressions and restricted nominal constructions from being overanalyzed as ordinary relative clauses. In particular, collocational, adjectival, copular, and construction-specific modifiers are resolved before the broader relative-clause-like classes. Instances that do not satisfy any of these conditions, including structurally inconsistent or mistagged cases, are assigned to \texttt{undefined}.

\section{Automatic annotation procedure}

{This section describes how the constructional typology introduced in Section~\ref{sec:typology} is operationalized over the KLUE dependency treebank.} The purpose of the procedure is not to define the typology computationally, but to apply the linguistic distinctions consistently to all extracted \texttt{ETM} instances. {The automatic annotation therefore serves as a reproducible implementation of the constructional analysis: it identifies the local morphosyntactic and lexical conditions under which a given \texttt{ETM} instance is assigned to one of the construction types.} This section specifies the annotation unit, the rule application order, the use of auxiliary lexical information, the treatment of unresolved cases, and the validation sample.

\paragraph{Annotation unit and extraction scope}

The annotation target is each \textit{eojeol} in the KLUE training data whose morphological analysis contains \texttt{ETM}. These instances are extracted automatically from the morphological layer of the corpus and paired with their local morphosyntactic context, including the internal morpheme sequence of the \textit{eojeol}, the immediately relevant neighboring morphemes, the following nominal head, and the local dependency configuration. {The extracted set constitutes the empirical domain over which the constructional typology is applied.}

Because the basic tokenization unit of KLUE is the \textit{eojeol}, a single annotation target may contain multiple morphemes and may realize a complex predicate sequence rather than a single lexical item. This is crucial for the present task, since the distinction among \texttt{ETM} constructions often depends on whether the adnominal form is associated with an adjective, a copular element, an auxiliary sequence, a bound noun, or a constructionally restricted head.

\paragraph{Rule inventory and ordered application}

Each extracted \texttt{ETM} instance is assigned a category by the ordered decision procedure defined in Algorithm~\ref{algo}. The procedure applies the symbolic conditions sequentially and returns the first matching label. {The ordering is linguistically motivated: constructionally specific patterns are resolved before broader relative-clause-like classes, because some surface forms that resemble relative clauses are better analyzed as fixed, restricted, or descriptive adnominal constructions.}

In practice, the procedure first tests collocational and non-relative construction-specific patterns, including adjectival and copular forms, and only afterwards the broader relative-clause-like classes. This precedence relation prevents expressions such as -에 대한 \textit{-e daehan} (`regarding') \(N\), \texttt{Verb+ETM} 때 \textit{ttae} (`time'), and 수 있다/없다 \textit{su itda/eopda} (`can/cannot') constructions from being overanalyzed as productive relative clauses. Thus, the rule order encodes the central claim of the typology: relative-clause-like modification is one constructional use of \texttt{ETM}, not the default interpretation of the ending.

\paragraph{Auxiliary lexical support}

The procedure primarily relies on morphosyntactic information available in KLUE itself. However, in selected cases it is supported by the auxiliary Sejong lexical resource described in Section~\ref{sec:data}. The resource is consulted when category assignment depends on lexical properties of the predicate that are not fully recoverable from the local surface pattern alone, especially with respect to predicate class and argument-structural compatibility.

The lexical resource is therefore used as a support layer rather than as an independent annotation source. {It helps determine whether a predicate--noun relation is compatible with a productive relative-clause-like analysis or whether another constructional interpretation is more appropriate.} The final labels are still assigned within the present typology and are determined by the ordered procedure.

\paragraph{Handling unresolved and inconsistent cases}

Not all extracted \texttt{ETM} instances can be assigned a linguistically interpretable label. When the source corpus contains morphological mistagging, structurally incomplete forms, or configurations that do not satisfy any of the defined conditions, the procedure assigns the residual label \texttt{undefined}. This prevents unreliable forced classification and makes the limits of the analysis explicit.

The \texttt{undefined} class therefore includes both source annotation errors and genuinely unresolved cases. These instances are retained in the released annotation layer for transparency, but they are distinguished from the linguistically interpreted classes in all downstream analysis.

\paragraph{Manual validation on a stratified sample}

To assess the reliability of the automatically derived \texttt{ETM} labels, we conducted a manual validation using category-stratified sampling from the final construction inventory. Cases assigned to the residual \texttt{undefined} category were excluded from the agreement analysis, since they do not correspond to an interpretable construction label in the proposed taxonomy. The resulting validation set contained 85 instances from the defined construction categories.

Each instance was presented with the source sentence, the target \textit{eojeol}, its internal morphological analysis, the following nominal head, the local syntactic context, and the automatically assigned label through a structured online form.\footnote{The manual validation interface is available at \url{https://forms.gle/5kNtyK7qnqdrJamn6}.} Two annotators independently judged whether the assigned label was appropriate in context, using the responses \textit{Yes}, \textit{No}, and \textit{Uncertain}. They gave identical judgments for 78 of the 85 instances, yielding an observed agreement of 91.8\% (95\% Wilson confidence interval: 84.0--96.0\%). Cohen's unweighted kappa, treating the three responses as nominal categories, was $\kappa=0.677$ (95\% percentile bootstrap confidence interval: 0.437--0.871; 200,000 resamples).

As shown in Table~\ref{tab:manual-validation-agreement}, the annotators jointly accepted the automatic label in 70 cases and jointly rejected it in 7 cases. They jointly selected \textit{Uncertain} for one case, while the remaining seven cases involved disagreement between a definite judgment and \textit{Uncertain}; no instance received conflicting \textit{Yes} and \textit{No} judgments. Among the 77 instances for which both annotators gave a definite judgment, the automatic label was accepted in 70 cases, or 90.9\% (95\% Wilson confidence interval: 82.4--95.5\%).

The jointly rejected cases reveal several recurrent sources of automatic-labeling error: predicate-class ambiguity, especially verbal forms misclassified as adjectival, as in 커진 \textit{keojin} `became larger'; confusion between ordinary nominal heads and bound or semantically light nouns, as in 점 \textit{jeom} `fact/point' and 예정 \textit{yejeong} `plan/schedule'; and construction-specific adnominal forms such as 다는 \textit{daneun}, which should not be collapsed with ordinary verbal relative modifiers. One additional rejected case, 알려진 \textit{allyeojin} `known/reported', reflects a residual boundary issue between a lexicalized passive verb analysis and a verb-plus-auxiliary/passive analysis. Disagreements and uncertain cases were concentrated mainly around the boundary between bound-noun constructions and productive relative-clause-like modifiers.

\begin{table}[!ht]
\centering
\footnotesize
\begin{tabular}{rr ccc}
\toprule
&& $\mathcal{A}_2$ & & \\
& & Yes & No & Uncertain \\
\midrule
$\mathcal{A}_1$ &Yes       & 70 & 0 & 6 \\
&No        & 0  & 7 & 1 \\
&Uncertain & 0  & 0 & 1 \\
\bottomrule
\end{tabular}
\caption{Cross-tabulation of the two annotators' judgments for the 85
manually validated instances.}
\label{tab:manual-validation-agreement}
\end{table}

\paragraph{Output representation}

The output of the procedure is an annotation layer that assigns one \texttt{ETM} construction type to every extracted target instance. Each annotation record retains the original sentence identifier, the target \textit{eojeol}, its assigned label, and the local structure needed for interpretation. {To respect the licensing conditions of the source treebank, we release the label schema, annotation code, documentation, and derived labels needed to reconstruct the annotation layer from the original KLUE data.} This representation makes it possible to analyze the distribution of \texttt{ETM} constructions while keeping the annotation procedure transparent and reproducible.

\section{Corpus analysis}

{This section presents the empirical distribution of \texttt{ETM} construction types in the training split of the KLUE dependency treebank.} The goal is not only to report raw frequencies, but also to show how strongly a coarse treatment of \texttt{ETM} obscures distinctions among relative-clause-like uses, non-relative descriptive modifiers, and constructionally restricted patterns.

\paragraph{Overall distribution of \texttt{ETM} types}

Table~\ref{etm-stat} reports the distribution of \texttt{ETM} categories in the KLUE training data.

\begin{table}[!ht]
\centering
\begin{tabular}{l r} \toprule
\textbf{Type} & \textbf{Count} \\ \midrule
\texttt{colloc+etm} & 563 \\
\texttt{modif+adj+etm} & 3394 \\
\texttt{modif+etm+ttae} & 119 \\
\texttt{modif+etm+nnb} & 2774 \\
\texttt{modif+ndaneun+etm} & 68 \\
\texttt{modif+su-iss+etm} & 65 \\
\texttt{modif+vcp+etm} & 842 \\
\texttt{modif+vx+vx+etm} & 3 \\
\texttt{rc+vv+etm} & 4816 \\
\texttt{rc+vv+vx+etm} & 293 \\
\texttt{rc+va+vx+etm} & 26 \\
\texttt{undefined} & 83 \\
\midrule
Total & 13,046 \\ \bottomrule
\end{tabular}
\caption{Distribution of \texttt{ETM} types in the KLUE training data.}
\label{etm-stat}
\end{table}

{The largest class is \texttt{rc+vv+etm} (4,816), showing that verbal relative-clause-like modification is a central use of \texttt{ETM}. It is not, however, the majority pattern.} The three relative-clause-like classes together account for 5,135 instances, or 39.4\% of the data. By contrast, non-relative and construction-specific classes account for 7,828 instances, or 60.0\%, with a further 83 instances, or 0.6\%, assigned to \texttt{undefined}.

{The main non-relative classes are \texttt{modif+adj+etm} (3,394) and \texttt{modif+etm+nnb} (2,774), followed by \texttt{modif+vcp+etm} (842) and \texttt{colloc+etm} (563).} These are not peripheral exceptions to a relative-clause pattern. They constitute a large part of the empirical distribution of \texttt{ETM}. Smaller categories, such as \texttt{modif+etm+ttae}, \texttt{modif+ndaneun+etm}, \texttt{modif+su-iss+etm}, and auxiliary-based relative-clause-like forms, further show that the distribution is shaped by specific constructional templates as well as by the broad contrast between relative and non-relative modification.

{The overall pattern is therefore heterogeneous. Treating every \texttt{ETM} instance as a relative clause marker would misclassify most tokens in the training data, conflating productive verbal modification with adjectival, copular, bound-nominal, modal, temporal, and collocational constructions.}

\paragraph{Relative clause and non-relative uses}

{The central quantitative question is how much of the \texttt{ETM} distribution is actually relative-clause-like.} To address this question, we aggregate the fine-grained labels into broader functional groups: relative-clause-like classes, non-relative descriptive modifiers, construction-specific modifiers, and unresolved cases.

{The relative-clause-like classes account for 5,135 of the 13,046 \texttt{ETM} tokens in the KLUE training data, or 39.4\% of the full distribution.} This group is dominated by \texttt{rc+vv+etm}, while \texttt{rc+vv+vx+etm} and \texttt{rc+va+vx+etm} contribute a smaller but structurally informative set of auxiliary-mediated cases. {The remainder of the distribution points in the opposite direction: 7,828 tokens, or 60.0\%, belong to non-relative or construction-specific classes, and a further 83 tokens, or 0.6\%, remain unresolved.} The quantitative balance therefore shows that relative-clause-like uses form an important subset of \texttt{ETM}, but not its corpus-level default.

{This asymmetry has a direct consequence for corpus identification. If every \texttt{ETM} token were treated by default as a relative clause marker, 7,911 of the 13,046 instances would be overgenerated, corresponding to 60.6\% of the data. Even if the unresolved cases are excluded and only resolved instances are considered, the overgeneration rate remains 60.4\%.} A morphology-based strategy would therefore misclassify most \texttt{ETM} occurrences, collapsing productive relativization into a broader domain that also includes adjectival modification, copular nominal modification, bound-nominal constructions, and lexicalized or constructionally restricted patterns.

{This comparison shows that the morphological presence of \texttt{ETM} is not, by itself, a reliable indicator of relative clause structure.}

\paragraph{Distribution by construction type and head noun}

{The distribution also shows that \texttt{ETM} constructions differ in the type of nominal head with which they combine.} Some classes are associated with relatively open nominal heads, while others are tied to fixed or highly restricted heads such as 때 \textit{ttae} (`time'), 수 \textit{su} (`possibility/way'), and content nouns such as 말 \textit{mal} (`word/speech'), 결정 \textit{gyeoljeong} (`decision'), and related nominal expressions.

The construction-specific categories are not distributed uniformly across nominal heads. Rather, they concentrate around a small number of recurrent head nouns and head-noun classes. The clearest cases are \texttt{modif+etm+ttae}, which is tied to the temporal bound noun 때 \textit{ttae} (`time'), and \texttt{modif+su-iss+etm}, which is anchored by the modal possibility construction built on 수 \textit{su} (`possibility/way'). The broader class \texttt{modif+etm+nnb} likewise clusters around dependent nominal heads, including bound nouns and semantically light content nouns that license recurrent modifier patterns rather than fully open relative clause formation.

A similar structural concentration appears in other parts of the inventory. \texttt{modif+vcp+etm} is associated with copular predication, where the adnominal form modifies a following nominal through identificational or classificatory structure rather than through verbal relativization. \texttt{colloc+etm}, by contrast, groups cases in which the modifier--head relation is lexically entrenched or strongly conventionalized. These patterns differ from \texttt{rc+vv+etm}, where the nominal head remains comparatively open and the modifier is more plausibly analyzed as a productive relative-clause-like predicate.

{The distribution therefore suggests that a substantial portion of \texttt{ETM} variation is organized around recurrent nominal anchors.} Some classes are defined by restricted head nouns, some by copular structure, and some by lexicalized modifier--noun pairings. {This is why the same overt ending appears across superficially similar but grammatically distinct configurations: the variation does not arise from free alternation, but from the interaction between adnominal morphology, predicate type, and construction-specific nominal environments.}

\begin{table}[!ht]
\centering
\begin{tabular}{p{0.23\linewidth} p{0.67\linewidth}}
\toprule
Category & Representative head noun type \\
\midrule
\texttt{rc+vv+etm}
& open lexical nouns \\

\texttt{rc+vv+vx+etm}
& open lexical nouns with auxiliary-mediated predicates \\

\texttt{modif+adj+etm}
& descriptive nominal heads \\

\texttt{modif+vcp+etm}
& nominal heads in identificational or classificatory constructions \\

\texttt{modif+etm+nnb}
& bound nouns and semantically light nominals, especially 것 \textit{geos} (`thing') \\

\texttt{modif+etm+ttae}
& 때 \textit{ttae} (`time') \\

\texttt{modif+su-iss+etm}
& nominal heads following the 수 \textit{su} (`possibility/way') modal construction \\

\texttt{modif+ndaneun+etm}
& content nouns such as 말 \textit{mal} (`word/speech') and 결정 \textit{gyeoljeong} (`decision') \\

\texttt{colloc+etm}
& nominal heads occurring in lexicalized or conventionalized adnominal expressions \\
\bottomrule
\end{tabular}
\caption{Representative head noun types associated with major \texttt{ETM} categories.}
\label{tab:etm-head-types}
\end{table}

{This head-noun sensitivity reinforces the main constructional claim: \texttt{ETM} does not define a uniform relative-clause structure, but participates in several adnominal constructions with different lexical and syntactic profiles.}

\paragraph{Problematic and undefined cases}

Finally, we examine the residual \texttt{undefined} class and other borderline cases. Although this class is numerically small, {83 instances, or 0.6\% of the data,} it is analytically important because it reveals the limits of both the source corpus and the current typology. The most common sources of failure include source mistagging, missing copular morphology, and structurally incomplete forms.

The \texttt{undefined} cases fall into a small number of recurrent subtypes rather than forming a homogeneous remainder. One subtype consists of tokens whose local morphology is inconsistent with the syntactic environment, typically because the source annotation assigns \texttt{ETM} where a different functional analysis would be required. A second subtype involves sequences in which a copular or auxiliary element appears to be missing from the annotated string, leaving an adnominal form that cannot be classified reliably as either a productive relative-clause-like modifier or a non-relative construction. A third subtype includes structurally incomplete forms, such as truncated expressions, fragmentary modifiers, or cases in which the nominal head or the relevant predicate sequence is not sufficiently represented for stable interpretation. There are also a small number of genuinely borderline configurations in which more than one analysis remains plausible even after local rule application.

{These residual cases arise primarily from limitations in the source annotation rather than from systematic gaps in the constructional typology itself.} Most reflect annotation error or representational underspecification in KLUE, especially where morphological segmentation or category assignment obscures the presence of copular, auxiliary, or bound-nominal structure. A smaller subset involves morphologically incomplete or fragmentary tokens that do not provide enough information for constructional classification. Only a narrow residue appears to represent genuine grammatical ambiguity in context.

{The residual analysis clarifies that the typology does not merely partition well-behaved cases. It also exposes where source annotation and surface patterning are insufficient for reliable constructional interpretation.}

\section{Conclusion}

This paper argued that the Korean adnominal ending \texttt{ETM} should not be analyzed as a direct marker of relative-clause structure. Rather, \texttt{ETM} is a morphological form shared by several distinct adnominal constructions. To make this heterogeneity explicit, we proposed a corpus-based typology that distinguishes productive relative-clause-like uses from adjectival, copular, bound-nominal, modal, temporal, collocational, and other construction-specific patterns. We applied this typology to the KLUE dependency treebank through a reproducible annotation procedure. {The result is a construction-sensitive annotation layer that makes distinctions recoverable which are not directly available from the original dependency labels.}

The corpus analysis shows that relative-clause-like uses are central but not dominant in the KLUE training data. Many \texttt{ETM} instances belong to non-relative or constructionally restricted classes that would be conflated under a coarse modifier analysis. The theoretical consequence is that Korean relative clauses cannot be defined by adnominal morphology alone. The presence of \texttt{ETM} identifies an adnominal environment, but the analysis of that environment as relative-clause-like depends on the relation between the predicate and the nominal head. Predicate type, auxiliary structure, head-noun restrictions, and lexicalized modifier--noun pairings therefore help distinguish productive relative-clause-like modification from other constructions with the same morphological marking. {From a corpus-linguistic perspective, this also shows that morphologically defined search criteria are insufficient for identifying construction types without an additional layer of linguistic classification.}

The resulting annotation layer provides a transparent basis for studying these distinctions in corpus data. We release the label schema, annotation code, documentation, and derived labels needed to reconstruct the annotation layer from the original KLUE data. {The manual validation indicates that the rule-based labels are reliable for the major construction types, while also identifying the boundaries where further refinement is needed.} The residual \texttt{undefined} cases show that some instances remain difficult because of source annotation inconsistencies, structurally incomplete forms, or borderline constructional status. Future work can test the typology on additional Korean treebanks and further examine the boundaries among relative-clause-like, bound-nominal, and constructionally restricted adnominal patterns.


\appendix

\section{Full typology of \texttt{ETM} constructions}
\label{app:typology}

This appendix provides descriptive documentation for the constructional typology of \texttt{ETM}. For each category, we give a fuller characterization together with representative corpus examples. {The purpose is not only to document the annotation labels, but also to make explicit the constructional distinctions that underlie the classification.} The main text presents the conceptual structure of the typology and the decision procedure; the appendix provides the empirical details needed to interpret the individual categories.

\subsection{\texttt{colloc+etm}}
\label{app:colloc-etm}

{This category covers cases in which the \texttt{ETM} form participates in a collocational or semi-fixed adnominal expression rather than in a freely generated relative-clause-like modifier.} Although the surface pattern resembles ordinary adnominal modification, the relation between the \texttt{ETM} form and the following noun is conventionalized. In the present study, collocational cases are identified with reference to the list in Figure~\ref{list-of-collocation}, based on \citet{geunseok:2009}.

\begin{figure}[!ht]
\centering
\footnotesize
\begin{tabular}{|p{\textwidth}@{}|} \hline
~\\
-로 인한 \textit{-lo inhan} (`caused by'), 
-로 치면 \textit{-lo chimyeon} (`in terms of'), 
-에 걸친 \textit{-e geolchin} (`spanning/across'), 
-에 관한 \textit{-e gwanhan} (`concerning'), 
-에 대한 \textit{-e daehan} (`about/regarding'), 
-에 따르면 \textit{-e ttareumyeon} (`according to'), 
-에 따른 \textit{-e ttareun} (`according to/resulting from'), 
-에 비하면 \textit{-e bihamyeon} (`compared with'), 
-에 의하면 \textit{-e uihamyeon} (`according to'), 
-에 의한 \textit{-e uihan} (`by/due to'), 
-에 준하는 \textit{-e junhaneun} (`equivalent to'), 
-에 즈음한 \textit{-e jeueumhan} (`around/at the time of'), 
-에게 비롯한 \textit{-ege birothan} (`originating from'), 
-와 같은 \textit{-wa gateun} (`such as/like'), 
-와 관련한 \textit{-wa gwallyeonhan} (`related to'), 
-와 다름없는 \textit{-wa dareumeomneun} (`no different from'), 
-을 둘러싼 \textit{-eul dulleossan} (`surrounding/concerning'), 
-을 망라한 \textit{-eul mangnahan} (`covering/encompassing'), 
-을 비롯한 \textit{-eul birothan} (`including/starting with'), 
-을 위한 \textit{-eul wihan} (`for'), 
-을 전후한 \textit{-eul jeonhuhan} (`around/before and after'), 
-을 통한 \textit{-eul tonghan} (`through/via'), 
-을 향한 \textit{-eul hyanghan} (`toward') \\
~\\
\hline 
\end{tabular}
\caption{List of collocations with \texttt{ETM} based on \citet{geunseok:2009}.} 
\label{list-of-collocation}
\end{figure}

{In \eqref{00246}, 따른 \textit{ttareun} is the \texttt{ETM} form of the verb 따르다 \textit{ttareuda} (`to follow'), but the sequence 판단에 따른 것 \textit{pandan-e ttareun geos} functions as a conventionalized adnominal expression meaning `what is based on the judgment' or `a result of the judgment'.}

\begin{exe}

\ex \label{00246} 
\glll 이번 [조치는]$_{\textsc{arg0}}$ 자물쇠가 다리의 안전을 위협한다는 [판단에]$_{\textsc{arg1}}$ {\colorbox{red!30}{따른}}$_{\textsc{etm}}$ {\colorbox{blue!30}{것이다.}}$_{\textsc{head}}$ \\
\textit{ibeon} \textit{jochineun} \textit{jamulswae-ga} \textit{dari-ui} \textit{anjeon-eul} \textit{wiheomhanda-neun} \textit{pandan-e} \textit{ttareun} \textit{geos-ida.} \\
this [measure--\textsc{top}\textsc{arg0}] lock-\textsc{nom} bridge-\textsc{gen} safety-\textsc{acc} threaten-\textsc{ec} [judgment-\textsc{ajt}\textsc{arg1}] {\colorbox{red!30}{following}}$_{\textsc{etm}}$ {\colorbox{blue!30}{thing-\textsc{cop}}}$_{\textsc{head}}$ \\
\trans `This measure is based on the judgment that the locks threaten the safety of the bridge.'   \hfill {\scriptsize\texttt{dev\_00246\_wikitree}}
	
\end{exe}

\begin{exe}
\ex \label{00050} 
\glll 강 [감독에]$_{\textsc{arg0}}$ {\colorbox{red!30}{대한}}$_{\textsc{etm}}$ 법원의 {\colorbox{blue!30}{영장실질심사는}}$_{\textsc{head}}$ 주말, 늦어도 내주 초 진행될 것으로 보인다. \\
\textit{gang} \textit{gamdok-e} \textit{daehan} \textit{beob-won-ui} \textit{yeongjangsiljilsimsa-neun} \textit{jumal,} \textit{neujeodo} \textit{naeju cho} \textit{jinhaengdoel} \textit{geos-euro} \textit{boinda.} \\
Kang director-\textsc{arg0} {\colorbox{red!30}{regarding}}$_{\textsc{etm}}$ court-\textsc{gen} {{\colorbox{blue!30}{warrant review}}}$_{\textsc{head}}$ weekend, {at the latest} {next week} early conducted be expected. \\
\trans `The court's warrant review regarding Director Kang is expected to be conducted over the weekend, or early next week at the latest.' \hfill {\scriptsize\texttt{dev\_00050\_wikitree}}

\end{exe}

{These examples illustrate cases in which the \texttt{ETM} form is part of a collocational adnominal construction rather than a productive relative-clause-like modifier.} Such instances are therefore classified as \texttt{colloc+etm}.

\subsection{\texttt{modif+adj+etm}}
\label{app:modif-adj-etm}

{This category includes adjectival adnominal constructions in which an adjective appears with \texttt{ETM} and directly modifies a following noun.} These are descriptive modifiers rather than relative-clause-like constructions. The modified noun is interpreted as the bearer of a property, and the adnominal form does not introduce the same predicate--head relation as a productive verbal relative-clause-like modifier.

\begin{exe}
\ex \label{00361} 
\glll 지난달 27일 유튜브에 공개된 해당 영상은 17일 기준 조회수 23만 회를 넘기며 {\colorbox{red!30}{많은}}$_{\textsc{etm}}$ {\colorbox{blue!30}{관심을}}$_{\textsc{head}}$ 받고 있다. \\
\textit{jinandal} \textit{27il} \textit{yutyube-e} \textit{gonggaedoen} \textit{haedang} \textit{yeongsang-eun} \textit{17il} \textit{gijun} \textit{joe-su} \textit{23man} \textit{hoe-reul} \textit{neomgimyeo} \textit{maneun} \textit{gwansim-eul} \textit{batgo} \textit{itda.} \\
{last month} 27th YouTube-\textsc{loc} released that video-\textsc{top} 17th {as of} views 230,000 {time-\textsc{acc}} surpassed {\colorbox{red!30}{many}}$_{\textsc{etm}}$ {\colorbox{blue!30}{interest-\textsc{acc}}}$_{\textsc{head}}$ receiving is.\\
\trans `The video released on YouTube on the 27th of last month has surpassed 230,000 views as of the 17th, receiving a lot of interest.' \hfill {\scriptsize\texttt{dev\_00361\_wikitree}}
\end{exe}

\begin{exe}
\ex \label{00346} 
\glll 국세청이 15일부터 인터넷 '연말정산간소화서비스' 운영을 시작한 가운데 오류가 [발생해]$_{\textsc{arg_0}}$ 홈페이지 [접속이]$_{\textsc{arg_1}}$ {\colorbox{red!30}{불가능한}}$_{\textsc{etm}}$ {\colorbox{blue!30}{상태다.}}$_{\textsc{head}}$ \\
\textit{gugsecheong-i} \textit{15ilbuteo} \textit{inteones} \textit{'yeonmaljeongsangansohwaseobiseu'} \textit{unyeong-eul} \textit{sijaghan} \textit{gaunde} \textit{olyu-ga} \textit{balsaenghae} \textit{hompeiji} \textit{jeobsog-i} \textit{bulganeunghan} \textit{sangtaeda.}\\
{National Tax Service-\textsc{nom}} 15th-from internet {Year-end Tax Settlement Simplification Service} operation-\textsc{acc} began while error-\textsc{nom} [occurred]$_{\textsc{arg_0}}$ website [access-\textsc{nom}]$_{\textsc{arg_1}}$ {\colorbox{red!30}{impossible}}$_{\textsc{etm}}$ {\colorbox{blue!30}{state-\textsc{cop}.}}$_{\textsc{head}}$\\
\trans `While the National Tax Service began operating the internet `Year-end Tax Settlement Simplification Service' on the 15th, an error occurred, making access to the website impossible.'  \hfill {\scriptsize\texttt{dev\_00346\_wikitree}}

\end{exe}

{These examples illustrate cases in which \texttt{ETM} realizes adjectival adnominal modification rather than productive relative-clause-like modification.} Such instances are therefore classified as \texttt{modif+adj+etm}.

\subsection{\texttt{modif+etm+ttae}}
\label{app:modif-etm-ttae}

{This category includes temporal adnominal constructions in which a \texttt{Verb+ETM} sequence modifies \textit{때}.} While the structure resembles a relative clause on the surface, the head noun is constructionally fixed and contributes temporal meaning. These cases are therefore treated as a distinct temporal construction rather than as productive relative-clause-like instances.

\begin{exe}

\ex \label{00430} \glll 고기$_{\textsc{arg0}}$ {\colorbox{red!30}{먹을}}$_{\textsc{etm}}$ {\colorbox{blue!30}{때}}$_{\textsc{head}}$ 찍어 먹는 기름장, 흔히 참기름에 소금, 후추 등을 넣는데요. \\
\textit{gogi} \textit{meogeul} \textit{ttae} \textit{jjigeo} \textit{meongneun} \textit{gireumjang}, \textit{heunhi} \textit{chamgireum-e} \textit{sogeum}, \textit{huchu} \textit{deung-eul} \textit{neonneundeyo.} \\
[meat-\textsc{arg0}] {\colorbox{red!30}{eat}}$_{\textsc{etm}}$ {\colorbox{blue!30}{time}}$_{\textsc{head}}$ dip eat {oil sauce,} usually {sesame oil} salt, pepper etc.-\textsc{acc} put. \\
\trans `Oil sauce for dipping and eating meat, usually made with sesame oil, salt, and pepper.'     \hfill {\scriptsize\texttt{dev\_00430\_wikitree}}
\end{exe}

\begin{exe}

\ex \label{00450}\glll 위버 시장은 소녀상 [건립을]$_{\textsc{arg0}}$ {\colorbox{red!30}{결정할}}$_{\textsc{etm}}$ {\colorbox{blue!30}{때}}$_{\textsc{head}}$ 시의원 5명 가운데 혼자 반대했던 인물이다. \\
\textit{wibeo} \textit{sijang-eun} \textit{sonyeosang} \textit{geonlib-eul} \textit{gyeoljeong-hal} \textit{ttae} \textit{siuiwon} \textit{5myeong} \textit{gaunde} \textit{honja} \textit{bandaehaetdeon} \textit{inmul-ida.} \\
Weaver mayor-\textsc{top} {girl statue} [construction-\textsc{acc}] {\colorbox{red!30}{decide}}$_{\textsc{etm}}$ {\colorbox{blue!30}{time}}$_{\textsc{head}}$ councilman {5 people} among alone opposed person-be. \\
\trans `Mayor Weaver was the only person among the five council members who opposed the decision to construct the girl statue.' \hfill {\scriptsize\texttt{dev\_00450\_wikitree}}

\end{exe}

{These examples illustrate \texttt{Verb+ETM} constructions with the fixed temporal head noun \textit{때}.} Such instances are therefore classified as \texttt{modif+etm+ttae} rather than as productive relative-clause-like modifiers.

\subsection{\texttt{modif+etm+nnb}}
\label{app:modif-etm-nnb}

{This category includes adnominal constructions in which an
\texttt{ETM}-marked predicate modifies a bound noun or semantically light
nominal head, most commonly {것} \textit{geos} (`thing').} The nominal
head does not denote an independently characterized entity but instead
nominalizes or packages the preceding proposition. This category covers
general bound-noun constructions not assigned to more specific categories
such as \texttt{modif+etm+ttae} or \texttt{modif+su-iss+etm}.

\begin{exe}
\ex \label{00139}
\glll 이들은 이번 [조치가]$_{\textsc{arg0}}$ NYT의 장기적 [수익구조를]$_{\textsc{arg1}}$ {\colorbox{red!30}{유지하려는}}$_{\textsc{etm}}$ {\colorbox{blue!30}{것이라고}}$_{\textsc{head}}$ 밝혔다. \\
\textit{ideul-eun} \textit{ibeon} \textit{jochi-ga} \textit{NYT-ui} \textit{janggijeok} \textit{suikgujo-reul} \textit{yuji-haryeoneun} \textit{geos-irago} \textit{balghyeotda.} \\
they-\textsc{top} this [measure-\textsc{nom}]$_{\textsc{arg0}}$
NYT-\textsc{gen} long-term [revenue structure-\textsc{acc}]$_{\textsc{arg1}}$
{\colorbox{red!30}{maintain-intend}}$_{\textsc{etm}}$
{\colorbox{blue!30}{thing-\textsc{cop}-\textsc{quot}}}$_{\textsc{head}}$
stated. \\
\trans `They stated that this measure was intended to maintain the NYT's
long-term revenue structure.'
\hfill {\scriptsize\texttt{dev\_00139\_wikitree}}
\end{exe}

{This example illustrates an \texttt{ETM} form followed by the bound noun
{것}, which packages the preceding proposition as the complement of
the reporting predicate.} Such instances are therefore classified as
\texttt{modif+etm+nnb} rather than as productive relative-clause-like
modifiers.

\subsection{\texttt{modif+ndaneun+etm}}
\label{app:modif-ndaneun-etm}

{This category includes content-noun constructions in which an adnominal form in \textit{는다는}/\textit{ㄴ다는} modifies a noun such as \textit{말}, \textit{결정}, or related nominals.} These constructions are not treated as ordinary relative clauses because the modified noun typically hosts proposition-like or quotative content rather than functioning as an open nominal head in a productive relative-clause-like relation.

\begin{exe}
\ex \label{00245} \glll 헌법재판소는 2014년 12월 19일 재판관 8(인용) {:} {1(기각)의} 의견으로, 피청구인 통합진보당을 [해산하고]$_{\textsc{arg0}}$ 그 소속 [국회의원은]$_{\textsc{arg1}}$ [의원직을]$_{\textsc{arg2}}$ {\colorbox{red!30}{상실한다는}}$_{\textsc{etm}}$ {\colorbox{blue!30}{결정을}}$_{\textsc{head}}$ 선고하였다. \\
\textit{heonbeopjaepanso-neun} \textit{2014nyeon} \textit{12wol} \textit{19il} \textit{jaepangwan} \textit{8(in-yong)} {:} \textit{1(gi-gak)-ui} \textit{uigyeon-euro}, \textit{picheongguin} \textit{tonghapjinbodang-eul} \textit{haesan-hago} \textit{geu} \textit{sosok} \textit{gukhoe-uiwon-eun} \textit{uiwonjik-eul} \textit{sangsilhanda-neun} \textit{gyeoljeong-eul} \textit{seongohae-tta.}\\
{Constitutional Court-\textsc{top}} {2014 year} {12 month} {19 day} justices 8(adopt) : 1(dismiss)-\textsc{gen} {opinion with,} respondent {Unified Progressive Party-\textsc{acc}} [dissolve-\textsc{arg0}] its affiliated {[National Assembly member-\textsc{top}]} {[parliamentary seat-\textsc{acc}]} {\colorbox{red!30}{lose}}$_{\textsc{etm}}$ {\colorbox{blue!30}{decision-\textsc{acc}}}$_{\textsc{head}}$ ruled. \\
\trans `On December 19, 2014, the Constitutional Court ruled by a vote of 8 (adopt) to 1 (dismiss) to dissolve the respondent Unified Progressive Party and to strip its affiliated National Assembly members of their parliamentary seats.' \hfill {\scriptsize\texttt{dev\_00245\_wikitree}}
\end{exe}

\begin{exe}
\ex \label{00021}\glll 국가의 내면 실상은 이러하나 가장 큰 [위기는]$_{\textsc{arg0}}$ 아무도 [모르게]$_{\textsc{arg1}}$ {\colorbox{red!30}{다가온다는}}$_{\textsc{etm}}$ {\colorbox{blue!30}{말과}}$_{\textsc{head}}$ 같이 우리 국민은 아직 잠들어 있습니다. \\
\textit{gukga-ui} \textit{naemyeon} \textit{silsang-eun} \textit{ireohana} \textit{gajang} \textit{keun} \textit{wigineun} \textit{amudo} \textit{moreuge} \textit{dagaondaneun} \textit{mal-gwa} \textit{gati} \textit{uri} \textit{gungmin-eun} \textit{ajik} \textit{jamdeureo} \textit{issseubnida.} \\
country-\textsc{gen} internal reality-\textsc{top} although most big [crisis-\textsc{arg0}] nobody [knowingly-\textsc{arg1}] {\colorbox{red!30}{approaches}}$_{\textsc{etm}}$ {{\colorbox{blue!30}{words with}}$_{\textsc{head}}$} like our people-\textsc{top} still asleep are. \\
\trans `Although this is the internal reality of the country, like the saying that the biggest crisis approaches without anyone knowing, our people are still asleep.'  \hfill {\scriptsize\texttt{dev\_00021\_wikitree}}

\end{exe}

{These examples illustrate \texttt{ETM} forms in content-like or quotative adnominal constructions.} Such instances are therefore classified as \texttt{modif+ndaneun+etm} rather than as productive relative-clause-like modifiers.

\subsection{\texttt{modif+su-iss+etm}}
\label{app:modif-su-iss-etm}

{This category includes modal adnominal constructions involving \textit{수} together with \textit{있다}/\textit{없다}.} Although these expressions are clause-like on the surface, they are treated as a separate construction because the pattern is built around a restricted nominal element and expresses possibility, ability, or availability rather than ordinary relative-clause-like modification.

\begin{exe}
\ex \label{00203} \glll 고등학교에서 가장 큰 추억을 남기러 간 수학여행에서 이젠 더 이상 그 친구들과 함께 [할수]$_{\textsc{arg0}}$ {\colorbox{red!30}{있는}}$_{\textsc{etm}}$ {\colorbox{blue!30}{시간을}}$_{\textsc{head}}$ 잃어버린 채 돌아왔습니다. \\
\textit{godeunghakgyo-eseo} \textit{gajang} \textit{keun} \textit{chueogeul} \textit{namgireo} \textit{gan} \textit{suhak-yeohaeng-eseo} \textit{ijen} \textit{deo} \textit{isang} \textit{geu} \textit{chingu-deul-gwa} \textit{hamkke} \textit{halsu} \textit{inneun} \textit{sigan-eul} \textit{ireobeorin} \textit{chae} \textit{dorawatseubnida.}\\
{high school-\textsc{loc}} most big memory-\textsc{acc} leave went {field trip-\textsc{loc}} now no longer those {friends with} together {can do-\textsc{arg0}} {{\colorbox{red!30}{able to}}$_{\textsc{etm}}$} {{\colorbox{blue!30}{time-\textsc{acc}}}}$_{\textsc{head}}$ lost state returned.\\
\trans `I returned from the field trip, where I went to make the best memories in high school, having lost the time I could no longer spend with those friends.' \hfill {\scriptsize\texttt{dev\_00203\_wikitree}}
\end{exe}

\begin{exe}

\ex \label{00227} \glll 시의 중재노력 덕분에 신씨가 권리금 인상을 요구하던 1차 임차인을 빼고 건물주와 직접 협상할 [수]$_{\textsc{arg0}}$ {\colorbox{red!30}{있는}}$_{\textsc{etm}}$ {\colorbox{blue!30}{길이}}$_{\textsc{head}}$ 열렸다. \\
\textit{si-ui} \textit{jungjaenoryeok} \textit{deokbune} \textit{sinssi-ga} \textit{gwolligeum} \textit{insang-eul} \textit{yogu-hadeon} \textit{ilcha} \textit{imcha-in-eul} \textit{ppaego} \textit{geonmulju-wa} \textit{jikjeop} \textit{hyeopsanghal} \textit{su} \textit{inneun} \textit{gil-i} \textit{yeollyeotda.} \\
city-\textsc{gen} {mediation efforts} {thanks to} Shin-\textsc{hon}-\textsc{nom} {key money} increase-\textsc{acc} demanded first tenant-\textsc{acc} excluding landlord-with directly negotiate [ability-\textsc{arg0}] {{\colorbox{red!30}{able-to}}$_{\textsc{etm}}$} {\colorbox{blue!30}{way-\textsc{nom}}}$_{\textsc{head}}$ opened. \\
\trans `Thanks to the city's mediation efforts, Shin was able to exclude the first tenant who demanded a key money increase and directly negotiate with the landlord, opening a way to do so.' \hfill {\scriptsize\texttt{dev\_00227\_wikitree}}

\end{exe}

{These examples illustrate \textit{수} plus \textit{있다}/\textit{없다} constructions modifying restricted or semantically light head nouns such as \textit{시간} and \textit{길}.} Such instances are therefore classified as \texttt{modif+su-iss+etm} rather than as productive relative-clause-like modifiers.

\subsection{\texttt{modif+vcp+etm}}
\label{app:modif-vcp-etm}

{This category covers copular and negative-copular adnominal constructions.} The modifier typically expresses identity, classification, role, or negated membership, as in \textit{운전자인 N} or \textit{아닌 N}. These patterns are treated separately from adjectival modifiers because the copular source and the resulting nominal interpretation are structurally distinct.

\begin{exe}

\ex \label{00144}\glll 이씨는 ... {\colorbox{red!30}{운전자인}}$_{\textsc{etm}}$ {\colorbox{blue!30}{김모(34)씨에게}}$_{\textsc{head}}$ 위협운전을 한 혐의를 받고 있다. \\
\textit{issi-neun} ... \textit{unjeonja-in} \textit{kimmo(34)-ssi-ege} \textit{wiheop-unjeon-eul} \textit{han} \textit{hyeomui-reul} \textit{batgo} \textit{itda.} \\
Lee-\textsc{top} ... {\colorbox{red!30}{driver}}$_{\textsc{etm}}$ {\colorbox{blue!30}{Kim (34)-Mr.-to}}$_{\textsc{head}}$ {threatening driving-\textsc{acc}} did suspected-\textsc{acc} receiving is. \\
\trans `Lee is suspected of threatening driving against Mr. Kim (34), the driver ...' \hfill {\scriptsize\texttt{dev\_00144\_wikitree}}

\end{exe}

\begin{exe}
\ex \label{01023} \glll 따라서 [가족이]${\textsc{arg0}}$ {\colorbox{red!30}{아닌}}$_{\textsc{etm}}$ 그룹 {\colorbox{blue!30}{구성원이라면}}$_{\textsc{head}}$ 화장실 사용이 불편할 수도 있는 상황입니다. \\
\textit{ttareseo} \textit{gajok-i} \textit{anin} \textit{gureup} \textit{guseongwon-iramyeon} \textit{hwajangsil} \textit{sayong-i} \textit{bulpyeonhal} \textit{sudo} \textit{inneun} \textit{sanghwang-imnida.} \\
therefore family-\textsc{nom} {\colorbox{red!30}{not}}$_{\textsc{etm}}$ group {\colorbox{blue!30}{member-if}}$_{\textsc{head}}$ bathroom use-\textsc{nom} uncomfortable can be situation-be. \\
\trans `Therefore, if you are a group member rather than a family, the use of the bathroom can be uncomfortable.'  \hfill {\scriptsize\texttt{dev\_01023\_airbnb}}
\end{exe}

{These examples illustrate copular or negative-copular adnominal constructions rather than productive relative-clause-like modifiers.} Such instances are therefore classified as \texttt{modif+vcp+etm}.

\subsection{\texttt{modif+vx+vx+etm}}
\label{app:modif-vx-vx-etm}

{This category includes restricted auxiliary-based adnominal constructions in which an auxiliary sequence contributes wish, condition, stance, or related meanings before an adnominal form.} These cases are infrequent in the corpus, but they are kept separate because their interpretation depends on a constructional auxiliary pattern rather than on ordinary relative-clause-like modification.

\begin{exe}
\ex \label{00260} \glll ... 더 이상 새 정부에 누가 되지 [않았으면]$_{\textsc{arg0}}$ {\colorbox{red!30}{하는}}$_{\textsc{etm}}$ {\colorbox{blue!30}{마음으로}}$_{\textsc{head}}$ 직을 사임하는 것입니다. \\
{...} \textit{deo} \textit{isang} \textit{sae} \textit{jeongbu-e} \textit{nuga} \textit{doeji} \textit{anatseumyeon} \textit{haneun} \textit{maeum-euro} \textit{jik-eul} \textit{saim-haneun} \textit{geos-imnida.} \\
{...} further {no longer} {new} {government to} harm become {if not-\textsc{arg0}} {{\colorbox{red!30}{wish}}$_{\textsc{etm}}$} {{\colorbox{blue!30}{with heart}}$_{\textsc{head}}$} position-\textsc{acc} resign be. \\
\trans `... I am resigning from my position with the wish that I no longer become a burden to the new government.' \hfill {\scriptsize\texttt{dev\_00260\_wikitree}}
\end{exe}

{This example illustrates an auxiliary-based \texttt{ETM} construction modifying a restricted noun such as \textit{마음}.} Such instances are therefore classified as \texttt{modif+vx+vx+etm} rather than as productive relative-clause-like modifiers.

\subsection{\texttt{rc+vv+etm}}
\label{app:rc-vv-etm}

{This category covers productive verbal relative-clause-like adnominal constructions.} The modifier is headed by a verbal predicate, including simple verbs and verbalized nouns with \texttt{XSV}, and the modified noun is interpreted as an open nominal head associated with the predicate.

\begin{exe}
\ex \label{00455} \glll [이른바]$_{\textsc{arg0}}$ '땅콩 리턴' [논란을]$_{\textsc{arg1}}$ {\colorbox{red!30}{빚은}}$_{\textsc{etm}}$ 조현아 대한항공 {\colorbox{blue!30}{부사장}}$_{\textsc{head}}$ 퇴진을 '20자'로 압축해 비판한 것이다. \\
\textit{ireunba} \textit{ttangkong} \textit{riteon} \textit{nollan-eul} \textit{bijeun} \textit{johyeona} \textit{daehan-hanggong} \textit{busajang} \textit{toejin-eul} \textit{20ja-ro} \textit{abchukhae} \textit{bipan-han} \textit{geos-ida.} \\
[so-called-\textsc{arg0}] 'nut return' [controversy-\textsc{acc}\textsc{arg1}] {\colorbox{red!30}{caused}}$_{\textsc{etm}}$ {Cho Hyun-ah} {Korean Air} {{\colorbox{blue!30}{vice president}}}$_{\textsc{head}}$ resignation-\textsc{acc} {`in 20 characters'-into} condensed criticized be. \\
\trans `It is a criticism condensed into 20 characters regarding the resignation of Korean Air Vice President Cho Hyun-ah, who caused the so-called 'nut return' controversy.' \hfill    {\scriptsize\texttt{dev\_00455\_wikitree}}
\end{exe}

\begin{exe}

\ex \label{00490} \glll [처음엔]$_{\textsc{arg0}}$ '이러다 말겠지'라 생각한 [듯]$_{\textsc{arg1}}$ 그저 [그렇게]$_{\textsc{arg2}}$ {\colorbox{red!30}{대하던}}$_{\textsc{etm}}$ 마을 주민들의 {\colorbox{blue!30}{태도가}}$_{\textsc{head}}$ 달라지기 시작했다. \\
\textit{cheoeum-en} \textit{ireoda} \textit{malgetji-ra} \textit{saenggakhan} \textit{deut} \textit{geujeo} \textit{geureoke} \textit{daehadeon} \textit{maeul} \textit{jumin-deul-ui} \textit{taedo-ga} \textit{dallajigi} \textit{sijakhaetda.} \\
[initially-\textsc{arg0}] 'it {will pass'-{}} thought [seemingly-\textsc{arg1}] just [so-\textsc{arg2}] {\colorbox{red!30}{treated}}$_{\textsc{etm}}$ village residents-\textsc{gen} {\colorbox{blue!30}{attitude-\textsc{nom}}}$_{\textsc{head}}$  change began. \\
\trans `Initially, the villagers, thinking 'this will pass,' just treated it as such, but their attitude began to change.' \hfill {\scriptsize\texttt{dev\_00490\_wikitree}}
\end{exe}

{These examples illustrate productive relative-clause-like modifiers in which a verbal predicate in \texttt{ETM} form modifies an open head noun.} Such instances are therefore classified as \texttt{rc+vv+etm}.

\subsection{\texttt{rc+vv+vx+etm}}
\label{app:rc-vv-vx-etm}

{This category includes productive relative-clause-like adnominal constructions headed by a verbal predicate plus an auxiliary complex.} The modified noun is interpreted as associated with the main verbal predicate rather than with the auxiliary element alone. These cases are separated from \texttt{rc+vv+etm} because the auxiliary structure is recurrent and structurally visible in the morphological decomposition.

\begin{exe}
\ex \label{00424} \glll 저보다 더 깊은 철학과 경륜과 뛰어난 인품을 겸비한 분께서 혁신 교육철학과 정책을 이어주시리라 믿으며 떨어지지$_{\textsc{vv}}$ {\colorbox{red!30}{않는}}$_{\textsc{etm}}$ {\colorbox{blue!30}{발걸음을}}$_{\textsc{head}}$ 옮기겠습니다. \\
\textit{jeo-boda} \textit{deo} \textit{gipeun} \textit{cheolhak-gwa} \textit{gyeongryun-gwa} \textit{ttwieonan} \textit{inpum-eul} \textit{gyeombihan} \textit{bun-kkeseo} \textit{hyeoksin} \textit{gyoyukcheolhak-gwa} \textit{jeongchaek-eul} \textit{ieojusirila} \textit{mideumyeo} \textit{tteoreojiji} \textit{anneun} \textit{balgeoreum-eul} \textit{omgigesseumnida.} \\
I-than more profound {philosophy-and} {experience-and} outstanding character-\textsc{acc} combine person innovation {education philosophy-and} policy–\textsc{acc} continue  believe fall\textsubscript{\textsc{vv}} {\colorbox{red!30}{not}}$_{\textsc{etm}}$ {\colorbox{blue!30}{steps-\textsc{acc}}}$_{\textsc{head}}$ move. \\
\trans `I believe that someone with deeper philosophy, greater experience, and outstanding character will continue the innovative educational philosophy and policies, and I will move forward with unwavering steps.'  \hfill {\scriptsize\texttt{dev\_00424\_wikitree}}
\end{exe}

\begin{exe}

\ex \label{00325} \glll 그러나 이미 영업이 끝난 시간인데다 건물 안에 [남아]$_{\textsc{vv}}$ {\colorbox{red!30}{있던}}$_{\textsc{etm}}$ {\colorbox{blue!30}{사람들이}}$_{\textsc{head}}$ 재빨리 대피해 인명피해는 없었다. \\
\textit{geureona} \textit{imi} \textit{yeongeobi} \textit{kkeutnan} \textit{sigan-indae-da} \textit{geonmul} \textit{an-e} \textit{nama} \textit{itdeon} \textit{saramdeuri} \textit{jaepalli} \textit{daepihae} \textit{inmyeongpihae-neun} \textit{eopseotda.} \\
however already business-\textsc{nom} ended time  building inside remained$_{\textsc{vv}}$ {\colorbox{red!30}{were}}$_{\textsc{etm}}$ {\colorbox{blue!30}{people-\textsc{nom}}}$_{\textsc{head}}$ quickly evacuated casualties-\textsc{top} none. \\
\trans `However, as it was already after business hours and the people remaining inside the building quickly evacuated, there were no casualties.'  \hfill {\scriptsize\texttt{dev\_00325\_wikitree}}

\end{exe}

{These examples illustrate productive relative-clause-like modifiers in which a verbal predicate plus auxiliary complex in \texttt{ETM} form modifies an open head noun.} Such instances are therefore classified as \texttt{rc+vv+vx+etm}.

\subsection{\texttt{rc+va+vx+etm}}
\label{app:rc-va-vx-etm}

{This category includes adjective-based relative-clause-like adnominal constructions involving an auxiliary complex.} These cases are structurally parallel to \texttt{rc+vv+vx+etm}, but the main descriptive element is adjectival rather than verbal.

\begin{exe}
\ex \label{01125} \glll 관광지들을 여행하기 나쁘지$_{\textsc{va}}$ {\colorbox{red!30}{않은}}$_{\textsc{etm}}$ {\colorbox{blue!30}{숙소입니다.}}$_{\textsc{head}}$ \\
\textit{gwangwangjideul-eul} \textit{yeohaenghagi} \textit{nappeuji} \textit{aneun} \textit{suksso-imnida.} \\
{tourist spots-\textsc{acc}} travel {not bad$_{\textsc{va}}$} {\colorbox{red!30}{not}}$_{\textsc{etm}}$ {\colorbox{blue!30}{accommodation-\textsc{cop}}}$_{\textsc{head}}$ \\
\trans `It is an accommodation not bad for traveling to tourist spots.' \hfill {\scriptsize\texttt{dev\_01125\_airbnb}}

\end{exe}

{This example illustrates a productive relative-clause-like modifier in which an adjectival predicate plus auxiliary complex in \texttt{ETM} form modifies an open head noun.} Such instances are therefore classified as \texttt{rc+va+vx+etm}.

\subsection{\texttt{undefined}}
\label{app:undefined-etm}

{This category is reserved for instances in which no reliable constructional analysis can be assigned.} The most common reasons are annotation error in the source corpus, incomplete structure, or mismatch between the morphological label and the actual form. These cases are retained for transparency, but they are excluded from the linguistically interpretable construction types.

\begin{exe}
\ex \label{00367} \glll 지난 4월 미스 온두라스에 선발된 {\colorbox{red!30}{알바라도는}}${\textsc{etm}}$ $\cdots$ 변을 {\colorbox{blue!30}{당했다.}}${\textsc{head}}$ \\
\textit{jinan} \textit{4-wol} \textit{miseu} \textit{onduraseu-e} \textit{seonbaldoen} \textit{albarado-neun} $\cdots$ \textit{byeon-eul} \textit{danghaetda.} \\
last April Miss Honduras-\textsc{dat} selected {\colorbox{red!30}{Alvarado$_{\textsc{top}}$}} $\cdots$ accident-\textsc{acc} {\colorbox{blue!30}{suffered}}. \\
\trans `Last April, Alvarado, who was selected as Miss Honduras, suffered an accident $\cdots$' \hfill {\scriptsize\texttt{dev\_00367\_wikitree}}
\end{exe}

\begin{exe}
\ex \label{02072} 
\glll 만약 리더가 $\cdots$ 사람이라면 밑의 사람들은 어떤 {\colorbox{red!30}{상황에서던}}$_{\textsc{etm}}$ 말 하지 않아도 그것을 최우선으로 {\colorbox{blue!30}{두고}}$_{\textsc{head}}$ 행동한다. \\
\textit{manyak} \textit{lideo-ga} $\cdots$ \textit{saram-iramyeon} \textit{mit-ui} \textit{saramdeul-eun} \textit{eotteon} \textit{sanghwang-eseo-deon} \textit{mal} \textit{haji} \textit{anado} \textit{geugeos-eul} \textit{choeuseon-euro} \textit{dugo} \textit{haengdonghanda.} \\
if leader-\textsc{nom} $\cdots$ person-if under people-\textsc{pl}-\textsc{top} any {\colorbox{red!30}{situation}}$_{\textsc{etm}}$ say not even that-\textsc{acc} {top priority-as} {\colorbox{blue!30}{place}}$_{\textsc{head}}$ act. \\
\trans `If the leader is $\cdots$, the subordinates will act with it as their top priority in any situation, even without being told.' \hfill {\scriptsize\texttt{train\_02072\_wikitree}}
\end{exe}

{These examples illustrate cases in which source annotation is morphologically inconsistent or structurally incomplete, preventing reliable assignment to any defined \texttt{ETM} construction type.} Such instances are therefore assigned to \texttt{undefined}.

\end{document}